\documentclass{article}
\usepackage[utf8]{inputenc}

\usepackage{fullpage}
\usepackage{array}
\usepackage{caption}
\usepackage{subcaption}
\usepackage{graphicx}
\usepackage{hyperref}
\usepackage{multirow}
\usepackage{amsfonts}
\usepackage{amsmath}
\usepackage{geometry}
\usepackage{amssymb}
\usepackage{float}
\usepackage{tikz}
\usetikzlibrary{mindmap}
\usetikzlibrary{shapes, snakes, arrows}
\usetikzlibrary{positioning,fit,calc}
\usepackage{amsthm}
\usepackage{comment}
\usepackage{authblk}
\usepackage{booktabs}

\theoremstyle{plain}

\theoremstyle{definition}

\theoremstyle{remark}

\usepackage{graphicx}
\usepackage{amsmath}
\usepackage{amssymb}
\usepackage{mathtools}
\usepackage{amsthm}
\usepackage{booktabs}
\usepackage{hyperref}
\usepackage{xcolor}
\usepackage{comment}
\usepackage{colortbl}
\usepackage{subfiles}
\usepackage{tikz}

\usepackage{array}
\usepackage{url}

\usepackage[
    backend=biber,
    style=numeric,
    hyperref = true,
    natbib = true, 
    maxbibnames=9,
    maxcitenames=1
  ]{biblatex}

\title{Dancing in the Shadows: Harnessing Ambiguity for Fairer Classifiers}
\author[1,2]{Ainhize Barrainkua}
\author[1,3,4]{Paula Gordaliza}
\author[1,5]{Jose A. Lozano}
\author[1,2,6]{Novi Quadrianto}

\affil[1]{Basque Center for Applied Mathematics (BCAM), Spain}
\affil[2]{Predictive Analytics Lab (PAL), University of Sussex, UK}
\affil[3]{Universidad Pública de Navarra (UPNA), Spain}
\affil[4]{Mark{3}Institute for Advanced Materials and Mathematics (INAMAT$^2$), Spain}
\affil[5]{University of the Basque Country (UPV/EHU), Spain}
\affil[6]{Monash University, Indonesia}

\begin{document}

\maketitle

\begin{abstract}
This paper introduces a novel approach to bolster algorithmic fairness in scenarios where sensitive information is only partially known. In particular, we propose to leverage instances with uncertain identity with regards to the sensitive attribute to train a conventional machine learning classifier.
The enhanced fairness observed in the final predictions of this classifier highlights the promising potential of prioritizing ambiguity (i.e., non-normativity) as a means to improve fairness guarantees in real-world classification tasks.

\end{abstract}

\section{Introduction}

Algorithmic systems, designed to streamline decision processes and enhance efficiency, have permeated virtually every aspect of our lives. From credit approvals to hiring decisions, from predictive policing to healthcare recommendations, algorithms wield significant influence. Yet, this influence is not neutral, and the consequences could be disproportionate for diverse communities. Subtle biases embedded in training data, the choices made during model development, and the very nature of algorithmic decision-making are some potential reasons for inequitable treatment of certain demographic groups, perpetuating and, in some instances, exacerbating societal disparities. Consider, for instance, the use of predictive policing algorithms, where certain communities are subjected to heightened surveillance based on historical crime data, perpetuating a cycle of over-policing \cite{ensign2018runaway}. Similarly, in hiring practices, algorithms may inadvertently favor certain demographics, leading to underrepresentation and reinforcing existing inequalities in the workplace \cite{datta2015automated, dastin2022amazon}. Therefore, it is crucial to acknowledge the inherent biases and disparities that have emerged within these systems and propose innovative solutions to enhance their fairness guarantees.

It is already well-known that the naïve strategy of simply removing the sensitive information (e.g., gender, race, age, sexual orientation) of the instances does not improve the fairness guarantees of the classifier. This happens because there are several 'objective' attributes that are highly correlated with the sensitive attribute, and therefore, they encode redundant information \cite{hajian2012methodology, kilbertus2017avoiding, besse2022survey}. As an alternative, numerous fairness-enhancing interventions have been proposed, categorized as pre-, in-, and post-processing methods based on the stage at which fairness requirements are integrated. Pre-processing methods aim to mitigate bias in the dataset, in-processing techniques adjust classifier learning to incorporate fairness guarantees, and post-processing methods alter algorithmic predictions to ensure fair final outcomes. Refer to \cite{pessach2023algorithmic} for a comprehensive review on fairness-enhancing interventions. 

Conventional fairness-enhancing interventions can only operate under complete knowledge of the sensitive information. However, such assumption is unrealistic: in many real-world applications, obtaining the sensitive attribute value of all the instances is either illegal or impractical, as individuals may be reluctant to provide such information due to privacy concerns or fear of discrimination. This constraint underscores the importance of developing fair and effective algorithms that can operate with only partial access to sensitive attributes, without compromising privacy or perpetuating biases. 

Numerous methods operating within these conditions utilize an attribute classifier to forecast the sensitive information linked with each instance \cite{awasthi2021evaluating, kallus2022assessing,  diana2022multiaccurate, zhang2022correct, nam2022spread, lokhande2022towards, liang2023fair}. Afterwards, they select instances for which confident predictions were made and utilize them to train a traditional fairness-enhancing intervention. In our study, we also advocate for the use of attribute classifiers, albeit with a distinct (opposite) purpose. Unlike conventional methods that depend on clear-cut predictions—a practice that has sparked ethical apprehensions— we embrace those characterized by high uncertainty. This decision stems from our hypothesis that leveraging instances with high ambiguity in the sensitive information, or what we term \textit{non-normative} cases, holds the potential to improve the fairness guarantees of the final classifier that will be blind to the sensitive information.
%In our exploration of algorithmic fairness, we find inspiration in the diversity of perspectives and experiences present in non-normative individuals and communities. 
%Embracing the richness of diversity not only aligns with ethical imperatives but also serves as a guiding principle in our approach. 
By embracing the richness of diversity and the inclusion of non-normative instances in our training data, we aim to develop classification rules that transcend traditional norms and promotes equitable outcomes for all individuals, regardless of their sensitive attributes.

Experiments conducted on benchmark real-world high-stake classification tasks demonstrate that incorporating non-normativity indeed enhances the fairness guarantees of conventional machine learning (ML) classifiers. Moreover, these findings indicate that these fairness assurances are on par with those provided by popular fairness-enhancing interventions that presume complete knowledge of sensitive information, whereas our method only requires partial knowledge of such information.

\section{Notation and Theoretical Background}

\subsection{Notation and Problem Setting}

We consider a dataset consisting of both non-sensitive and sensitive information, denoted as $\mathcal{D}_1 = \{ (x_i, s_i) \}_{i=1}^N$, where $x\in \mathcal{X} \subset \mathbb{R}^d$ represents the non-sensitive features, and $s\in \mathcal{S}$ denotes the sensitive feature(s). This dataset will serve as the training data for the algorithm to detect instances with uncertain sensitive information. Additionally, we consider a second dataset containing non-sensitive features and class information, but without sensitive information, denoted as $\mathcal{D}_2 = \{ (x_i, y_i) \}_{i=1}^M$, where $y\in \mathcal{Y}$ represents the class label. Non-normative instances will be identified from this dataset and utilized to train the final classification rule, which will be blind to the sensitive information. For simplicity, we focus on binary classification tasks and binary sensitive attributes, where $\mathcal{Y}= \{ 0,1 \}$ and $\mathcal{S}= \{ 0,1 \}$, respectively. However, extending this work to multi-class and multi-group settings is straightforward.

\subsection{Measuring Algorithmic Fairness}

There are many definitions to measure the fairness guarantees of classifiers, and are conventionally divided into two groups: \textit{individual} and \textit{statistical} (or \textit{group}) notions. Individual fairness notions operate under the principle that comparable individuals should receive similar treatment. On the other hand, statistical fairness notions examine whether a particular statistical quantity is consistent across different sensitive groups. The primary limitations of individual notions include the necessity of defining a non-trivial similarity metric and the requirement to compare all potential pairs of instances, which can be impractical at scale. In contrast, statistical notions offer a more computationally efficient alternative, contributing to their widespread adoption in the literature. Within statistical notions, the name of the fairness metrics varies depending on the statistical quantity being considered. For instance, \textit{demographic parity} (DP) \cite{dwork2012fairness} assesses whether the acceptance rate is consistent across different sensitive groups, that is:
\begin{equation}
    \textrm{DP} = |p(\hat{y}=1|s=1)-p(\hat{y}=1|s=0)|.
\end{equation}
On the other hand, \textit{equality of opportunity} (EOp) \cite{hardt2016equality} examines whether the true positive rate (TPR) is uniform across these groups:
\begin{equation}
    \textrm{EOp} = |p(\hat{y}=1|s=1, y = 1)-p(\hat{y}=1|s=0, y = 1)|
\end{equation}

See \cite{pessach2023algorithmic} for a comprehensive review on fairness metrics and fairness-enhancing interventions.

\section{Methodology}

Our approach comprises three primary components: the norm breaker ensemble, the non-normative dataset, and the unbiased classifier. In the following lines, we present detailed explanations of each.

\paragraph{Norm Breaker Ensemble (NBE)} The primary objective of the NBE is to identify instances that deviate from the norm. NBE operates as a conventional ensemble classifier, flagging instances with elevated predictive uncertainty as less normative. Each base classifier within the ensemble serves as an attribute classifier ($h_S: \mathcal{X} \rightarrow \mathcal{S}$).
%capable of being any conventional machine learning classifier trained to predict the sensitive attributes of the instances. 
These attribute classifiers provide varying perspectives on the relationship between the 'objective' features and the sensitive attribute.
The ensemble is trained with the instances in $\mathcal{D}_1$, for which the sensitive information is known. 
The objective behind building such an ensemble is to detect non-normative instances. These instances are characterized by the highest ambiguity in the predicted sensitive attribute. In other words, they are the instances with significant disagreement in predictions among the base estimators of the ensemble. Specifically, each instance $(x, y)$ will be assigned an uncertainty value $u_x$, 
%reflecting the extent of non-normativity, 
calculated as follows: $u_x = 1 - \max \{ p(s=1|x), 1-p(s=1|x) \}$, where $p(s=1|x)$ refers to the probability given by the ensemble to assigning sensitive attribute value $s=1$ to instance $(x,y)$. Uncertainty reaches its highest value when the ensemble exhibit the greatest disagreement, meaning half of the learners support one value of the sensitive attribute while the remainder advocate for the opposite value.

\paragraph{Non-Normative Dataset ($\mathcal{D}'_2$)} The non-normative dataset comprises instances with uncertain or ambiguous sensitive information.
%Identify the non-normative instances. 
%We will consider two cases, one in which the demographics are known and one in which the sensitive information is unknown. 
Let $u_{x}$ be the predictive uncertainty of the NBE regarding the sensitive information of instance $(x,y)$, then the non-normative dataset is created as follows: $\mathcal{D}'_2 = \{ (x,y) \; | \; u_{x} \geq U \}$, where $U$ is the uncertainty threshold. 
Our approach targets the elimination of bias from the dataset utilized to train the final conventional ML classifier. Thus, it serves as a \textit{pre-processing} fairness-enhancing intervention.

\paragraph{Unbiased Classifier ($h_Y$)} The final unbiased classifier is also a conventional unconstrained ML classifier (blind to the sensitive information), which does not inherently include any explicit fairness-enhancing interventions (i.e., $h_Y : \mathcal{X} \rightarrow \mathcal{Y}$). This classifier will be trained using instances from $\mathcal{D}'_2$. Given that this dataset exhibits less bias than the original, it is expected that the ML classifier will generate less biased decisions. 
It is important to note that the family of functions selected to build this classifier may differ from the one chosen for the base learners of the NBE.

\section{Experiments}

\subsection{Experimental Setting}

\paragraph{Models} We explore different conventional ML classifiers: logistic regression (LR), support vector machine (SVM) and light gradient boosting method (LGBM) \cite{ke2017lightgbm}. These models serve as the base learners for the NBE, the final unbiased classifier and the classifer on top of which the state-of-the-art (SOTA) fairness-enhancing interventions are added. 
We construct the NBE using a particular family of functions by \textit{bagging} (a.k.a. boostrapping). 
Furthermore, the SOTA interventions considered are the pre-processing method by \citet{kamiran2012data} and the post-processing method by \citet{hardt2016equality}, respectively referenced as RW and PP throughout the remainder of the paper. Note that, unlike our approach, these fairness-enhancing interventions operate under the assumption of complete knowledge regarding the sensitive information. 

\paragraph{Datasets} (a) German Credit, data from a German bank to predict good/bad credit score, publicly available in the UCI repository \cite{asuncion2007uci}; (b) Adult Income, US census based data from 1994 for income prediction, publicly available in the UCI repository \cite{asuncion2007uci}; and (c) COMPAS \cite{angwin2022machine}, arrest data from Broward County, Florida,
originally compiled by ProPublica. Their corresponding details are outlined in Table \ref{tab:datasets}. 

\begin{table}[htbp]
\caption{Datasets: name, number of instances (size), number of features ($\#$feat), application domain and sensitive attribute (S).}
\begin{center}
\begin{tabular}{|c|c|c|c|c|}
\hline
\textbf{Dataset} & \textbf{Size} & \textbf{$\#$feat} & \textbf{Domain} & \textbf{S} \\
\hline
German Credit & 1,000 & 20 & Finance & Age \\
Adult Income & 48,842 & 15 & Finance & Race \\
COMPAS & 6,000 & 11 & Criminal justice & Race \\
%LSAC & 20,798 & 12 & Education & Gender \\
\hline
\end{tabular}
\label{tab:datasets}
\end{center}
\end{table}

\paragraph{Evaluation} We evaluate our approach for 10 different train/test (70/30) paritions and report averaged result in fairness guarantees (EOp and DP) and accuracy. We partition the training dataset into two disjoint subsets of equal size to form $\mathcal{D}_1$ and $\mathcal{D}_2$. In our method, $\mathcal{D}_1$ is used to train the attribute classifier and identify the non-normative instances from $\mathcal{D}_2$, resulting in the creation of $\mathcal{D}'_2$. On the other hand, the pre- and post-processing method considered are trained using the original dataset $\mathcal{D}_2$, but further assuming that the sensitive information of those instances is known. 
Additionally, we investigate the average uncertainty of the sensitive information in dataset $\mathcal{D}_2$ using NBE, as well as the accuracy of NBE in predicting the sensitive information of $\mathcal{D}_2$.

\subsection{Results}

\begin{figure*}[htbp]
    \centering
    \begin{subfigure}[b]{0.23\textwidth}
        \centering
        \caption{LR (EOp)}
        \includegraphics[width=\textwidth]{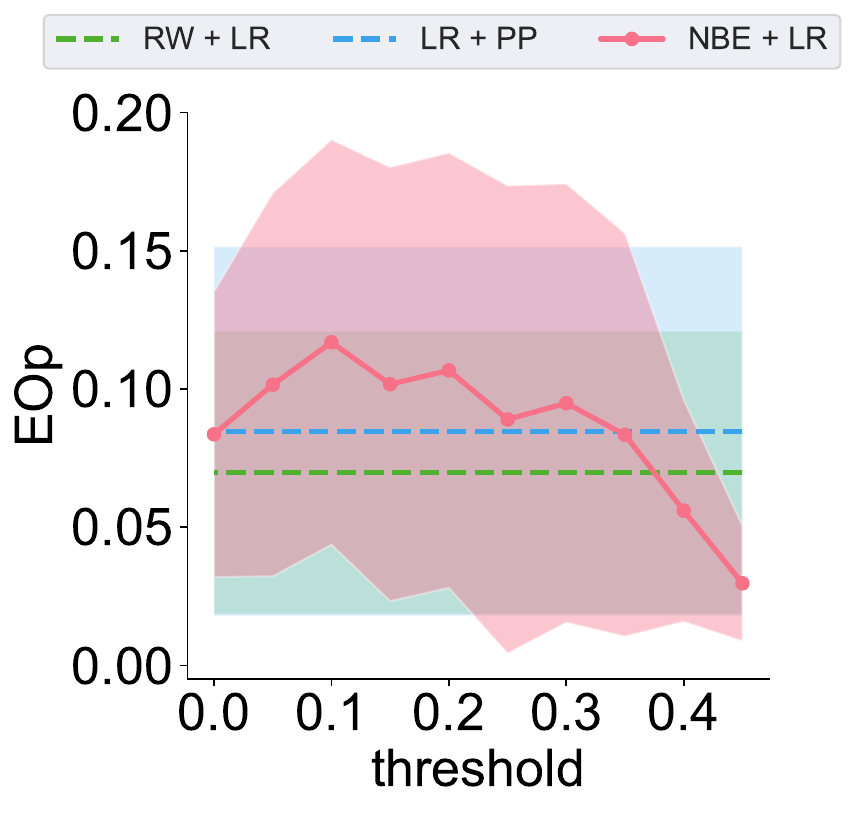}
        \label{fig:sub11g}
    \end{subfigure}
    \hfill
    \begin{subfigure}[b]{0.245\textwidth}
        \centering
        \caption{SVM (EOp)}
        \includegraphics[width=\textwidth]{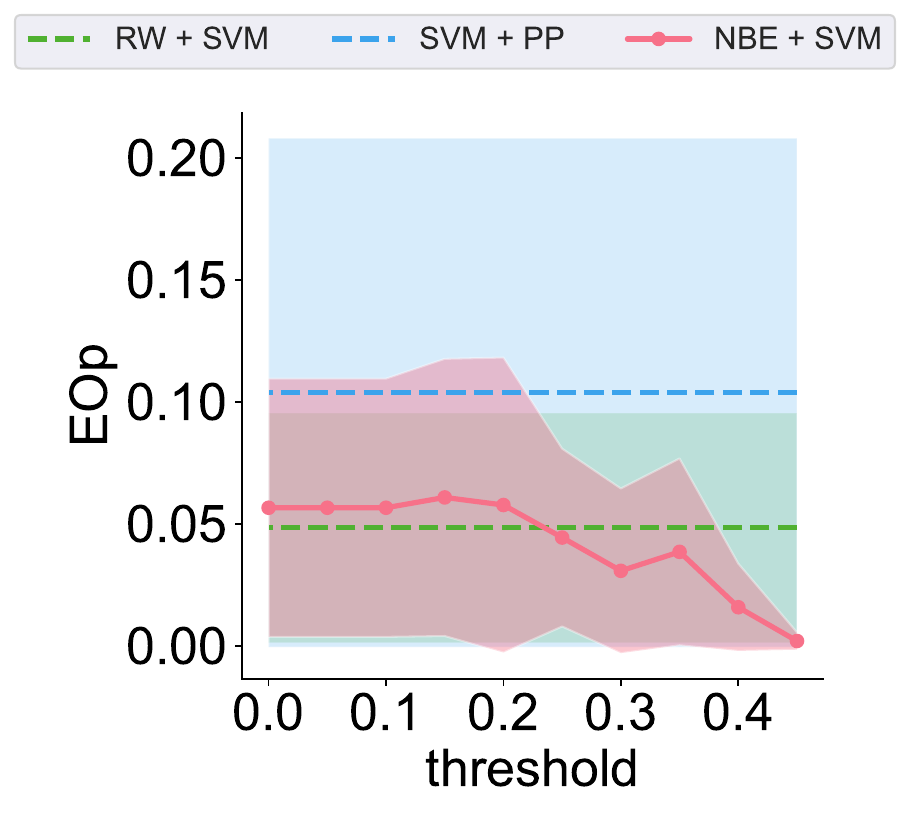}
        \label{fig:sub12g}
    \end{subfigure}
    \hfill
    \begin{subfigure}[b]{0.275\textwidth}
        \centering
        \caption{LGBM (EOp)}
        \includegraphics[width=\textwidth]{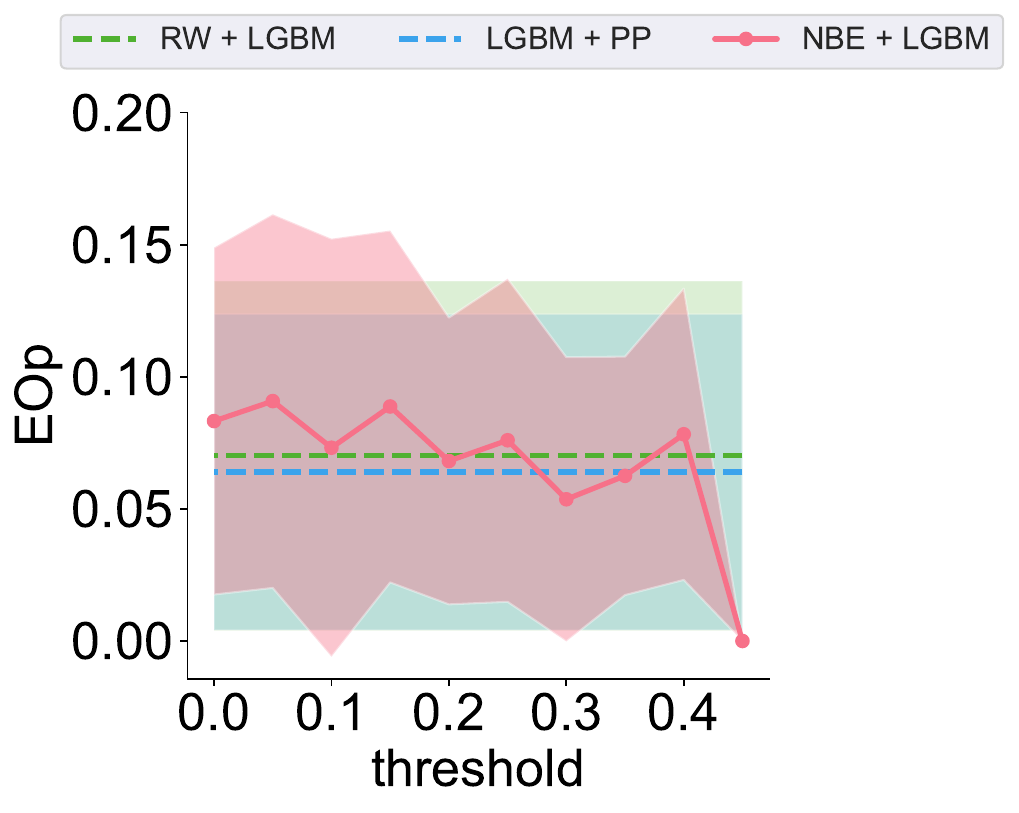}
        \label{fig:sub13g}
    \end{subfigure}
    \hfill 
    \\
    \begin{subfigure}[b]{0.23\textwidth}
        \centering
        \caption{LR (DP)}
        \includegraphics[width=\textwidth]{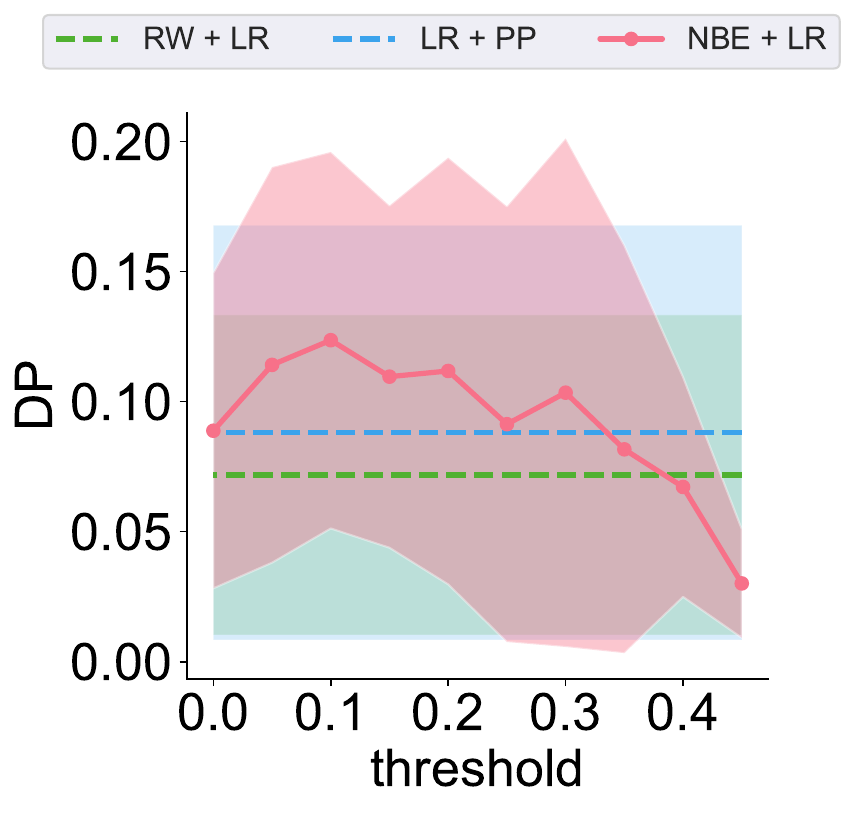}
        \label{fig:sub21g}
    \end{subfigure}
    \hfill
    \begin{subfigure}[b]{0.245\textwidth}
        \centering
        \caption{SVM (DP)}
        \includegraphics[width=\textwidth]{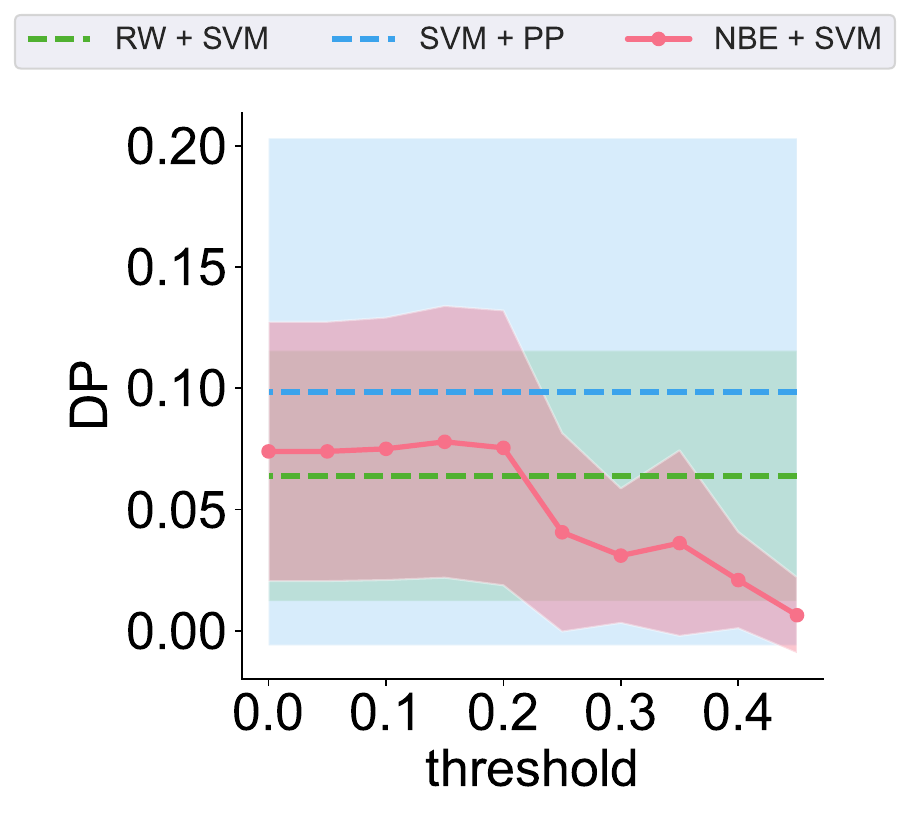}
        \label{fig:sub22g}
    \end{subfigure}
    \hfill
    \begin{subfigure}[b]{0.275\textwidth}
        \centering
        \caption{LGBM (DP)}
        \includegraphics[width=\textwidth]{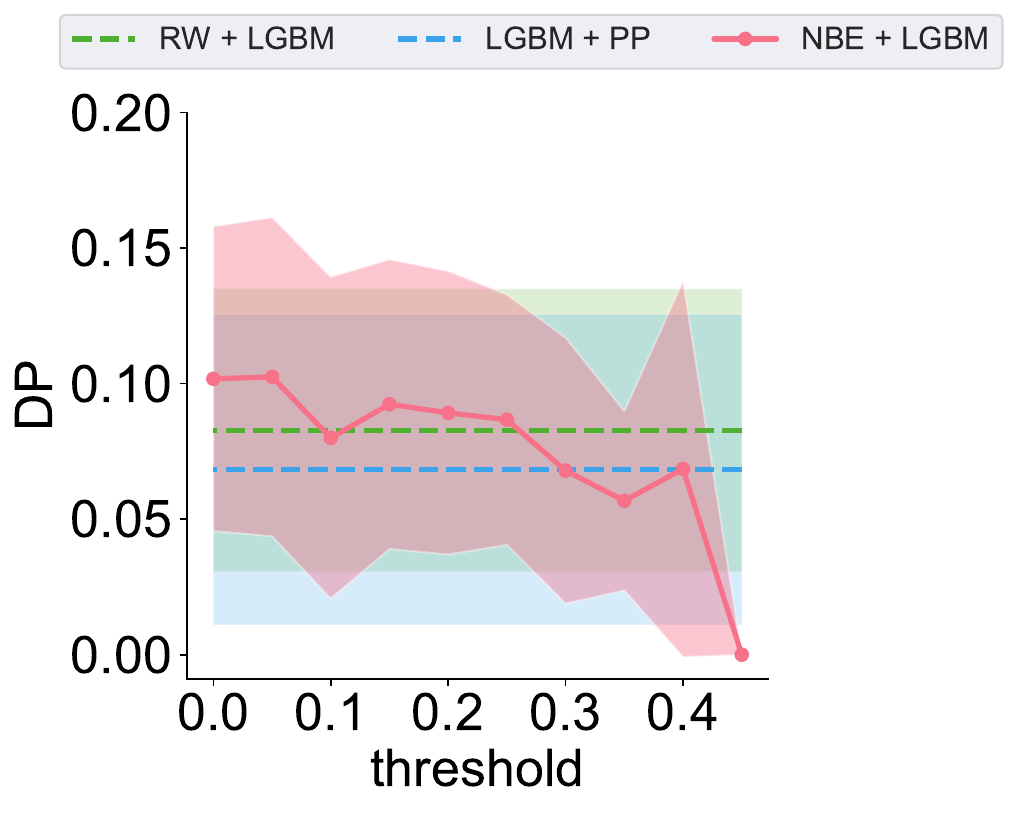}
        \label{fig:sub23g}
    \end{subfigure}
    \hfill
    \\
    \begin{subfigure}[b]{0.23\textwidth}
        \centering
        \caption{LR (EOp vs. acc)}
        \includegraphics[width=\textwidth]{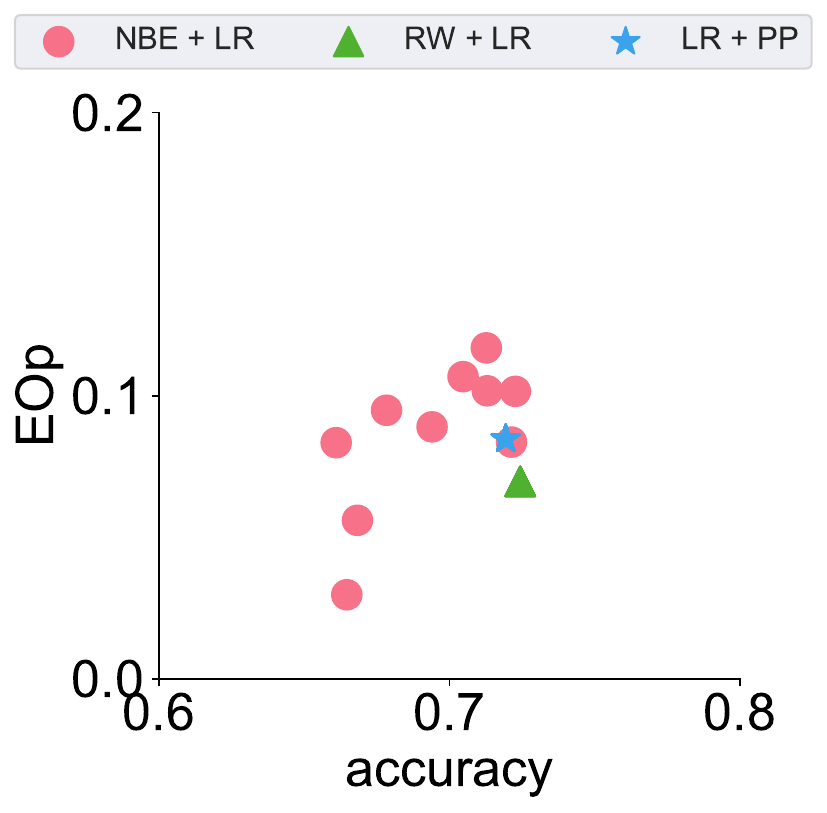}
        \label{fig:sub31g}
    \end{subfigure}
    \hfill
    \begin{subfigure}[b]{0.253\textwidth}
        \centering
        \caption{SVM (EOp vs. acc)}
        \includegraphics[width=\textwidth]{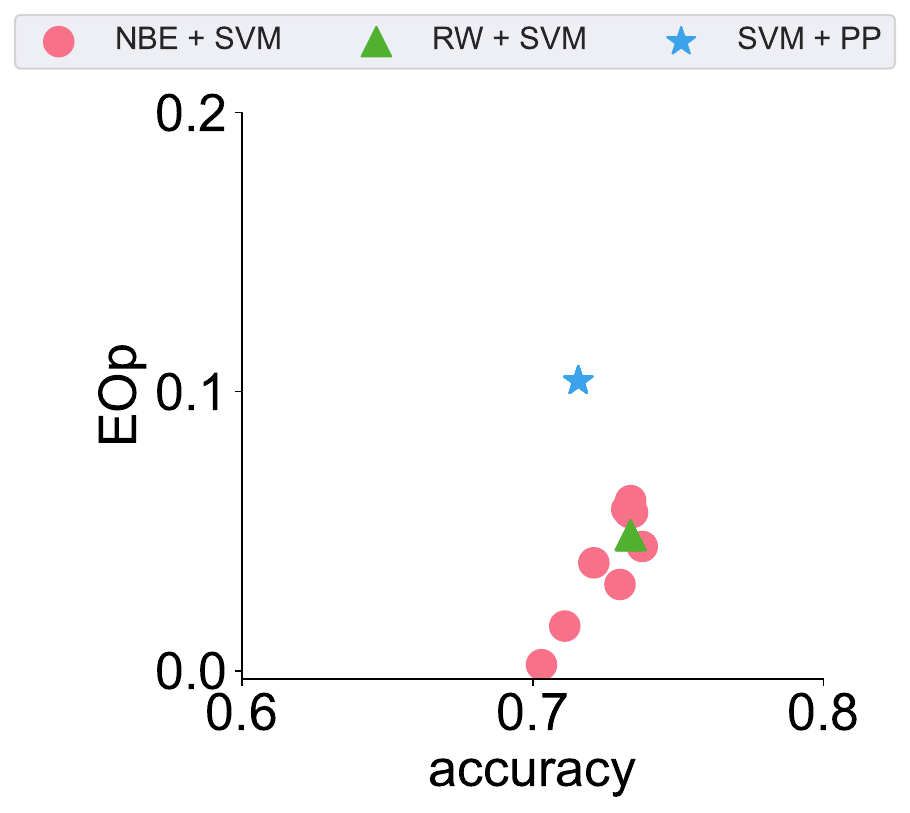}
        \label{fig:sub32g}
    \end{subfigure}
    \hfill
    \begin{subfigure}[b]{0.275\textwidth}
        \centering
        \caption{LGBM (EOp vs. acc)}
        \includegraphics[width=\textwidth]{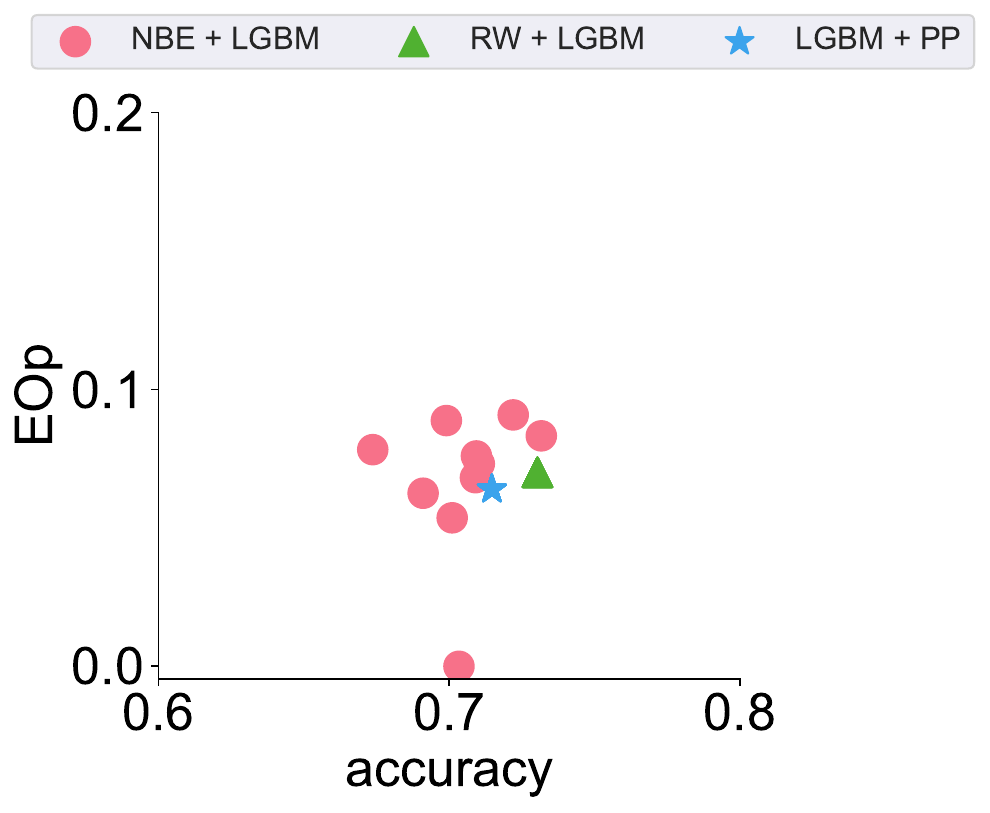}
        \label{fig:sub33g}
    \end{subfigure}
    \caption{ Average results in the German Credit dataset for varying uncertainty threshold values, using different base learners. The first two rows depict fairness guarantees in terms of EOp and DP, respectively, with the shaded region indicating variance. The last row illustrates the results in the joint space of fairness(EOp) and accuracy. Our method outperforms the SOTA interventions in terms of fairness across all ML classifiers. Specifically, SVM and LGBM offer the most optimal fairness guarantees, almost reaching perfection, with minimal compromise in accuracy. Moreover, SVM shows the best fairness-accuracy trade-off curve.
    %Three different base learners, (a) LR, (b) SVM, and (c) LGBM, are utilized for the NBE and the unbiased classifier. NBE + unbiased classifier is compared to popular pre- and post-processing fairness-enhancing interventions.
    }
    \label{fig:German}
\end{figure*}

\begin{figure*}[htbp]
    \centering
    \begin{subfigure}[b]{0.23\textwidth}
        \centering
        \caption{LR (EOp)}
        \includegraphics[width=\textwidth]{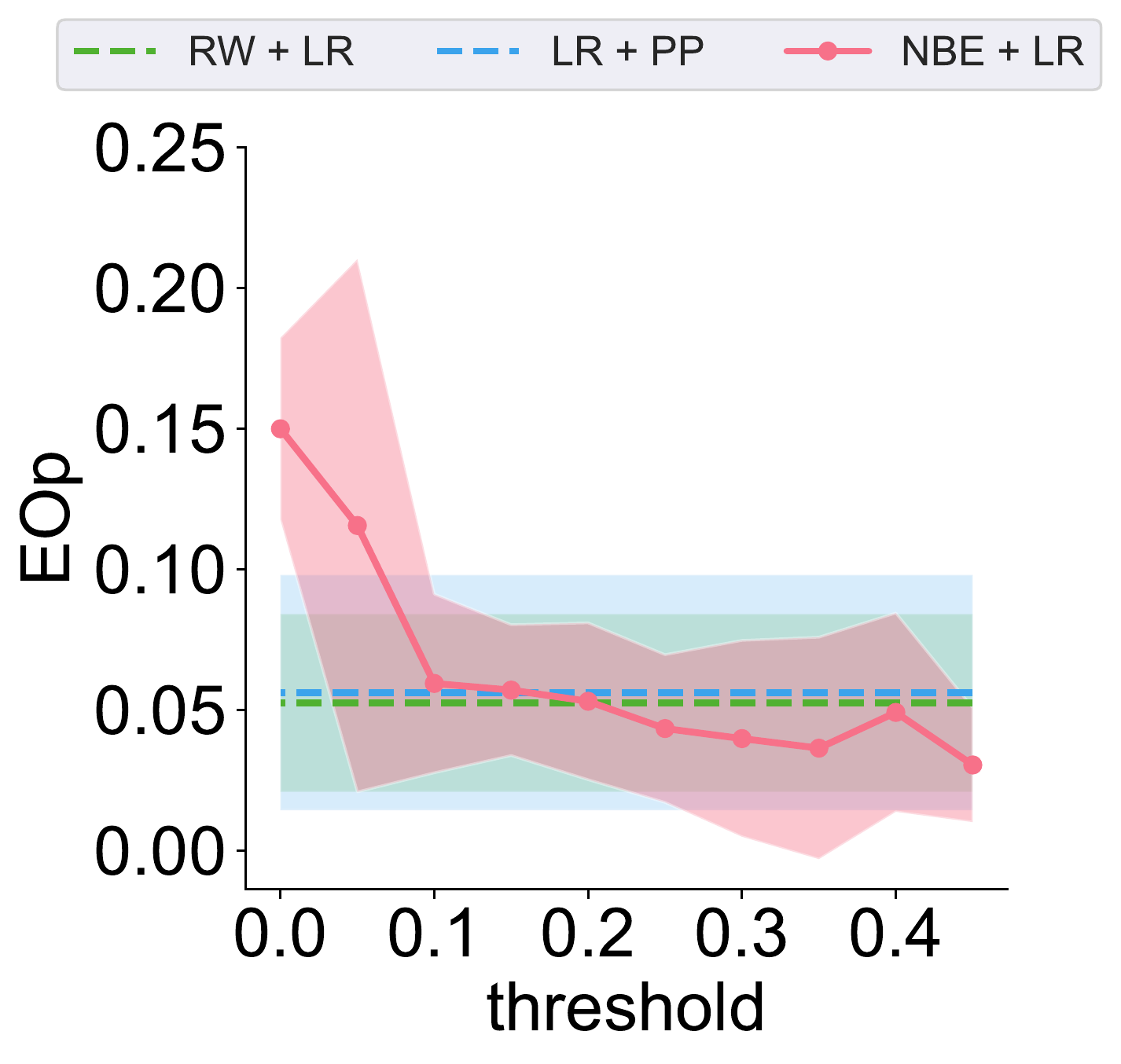}
        \label{fig:sub11a}
    \end{subfigure}
    \hfill
    \begin{subfigure}[b]{0.235\textwidth}
        \centering
        \caption{SVM (EOp)}
        \includegraphics[width=\textwidth]{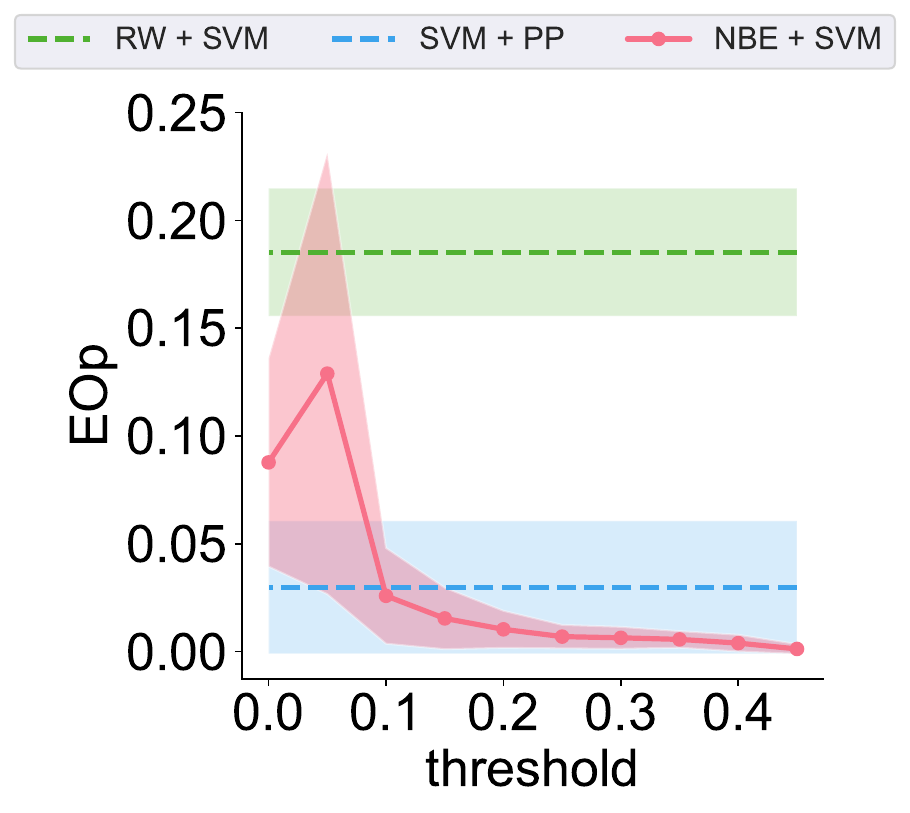}
        \label{fig:sub12a}
    \end{subfigure}
    \hfill
    \begin{subfigure}[b]{0.275\textwidth}
        \centering
        \caption{LGBM (EOp)}
        \includegraphics[width=\textwidth]{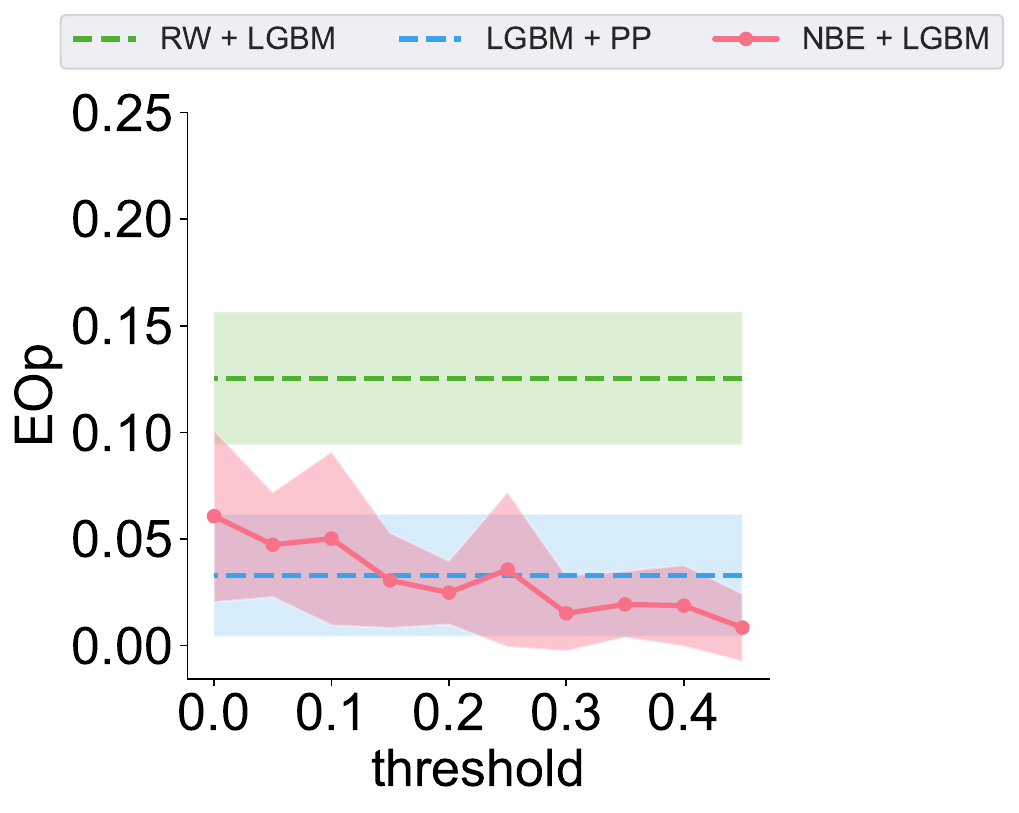}
        \label{fig:sub13a}
    \end{subfigure}
    \hfill
    \\
    \begin{subfigure}[b]{0.23\textwidth}
        \centering
        \caption{LR (DP)}
        \includegraphics[width=\textwidth]{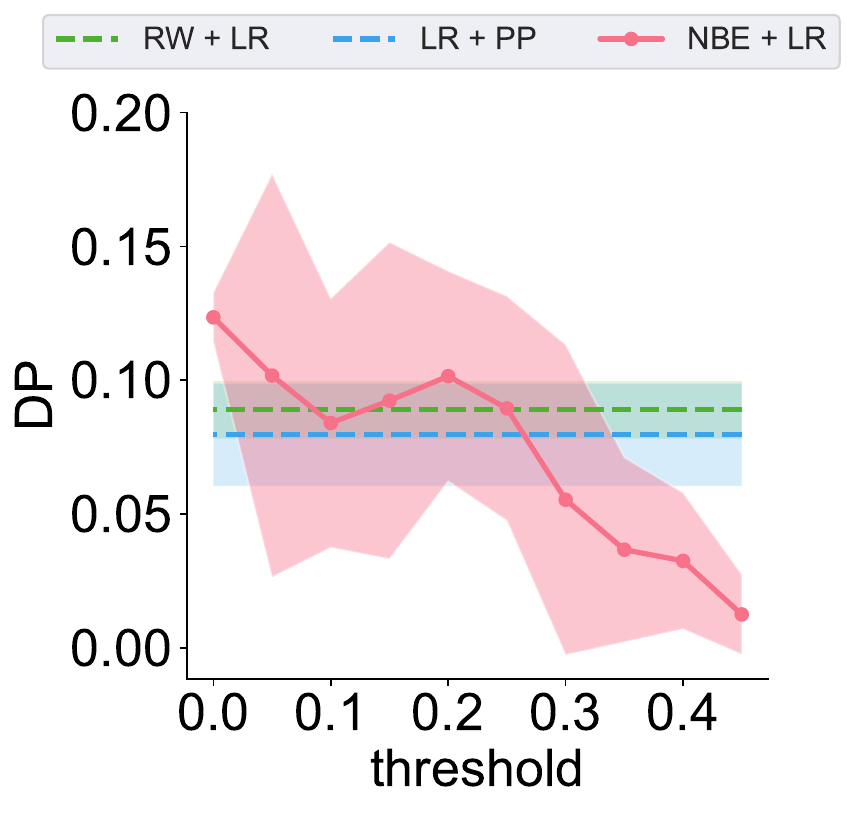}
        \label{fig:sub21a}
    \end{subfigure}
    \hfill
    \begin{subfigure}[b]{0.245\textwidth}
        \centering
        \caption{SVM (DP)}
        \includegraphics[width=\textwidth]{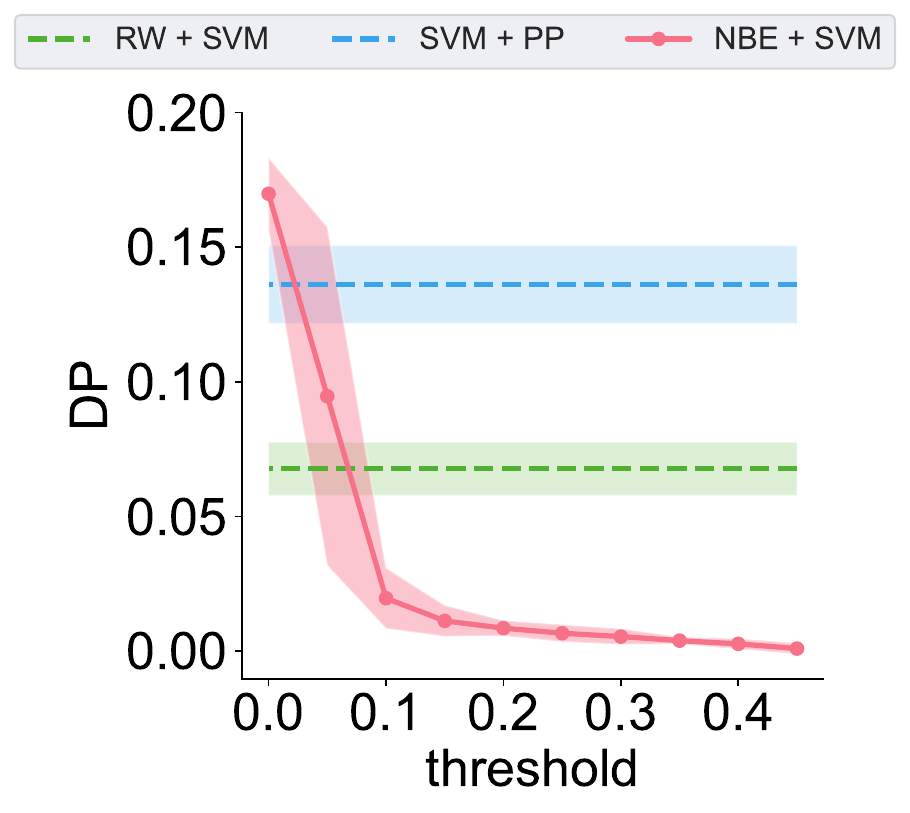}
        \label{fig:sub22a}
    \end{subfigure}
    \hfill
    \begin{subfigure}[b]{0.275\textwidth}
        \centering
        \caption{LGBM (DP)}
        \includegraphics[width=\textwidth]{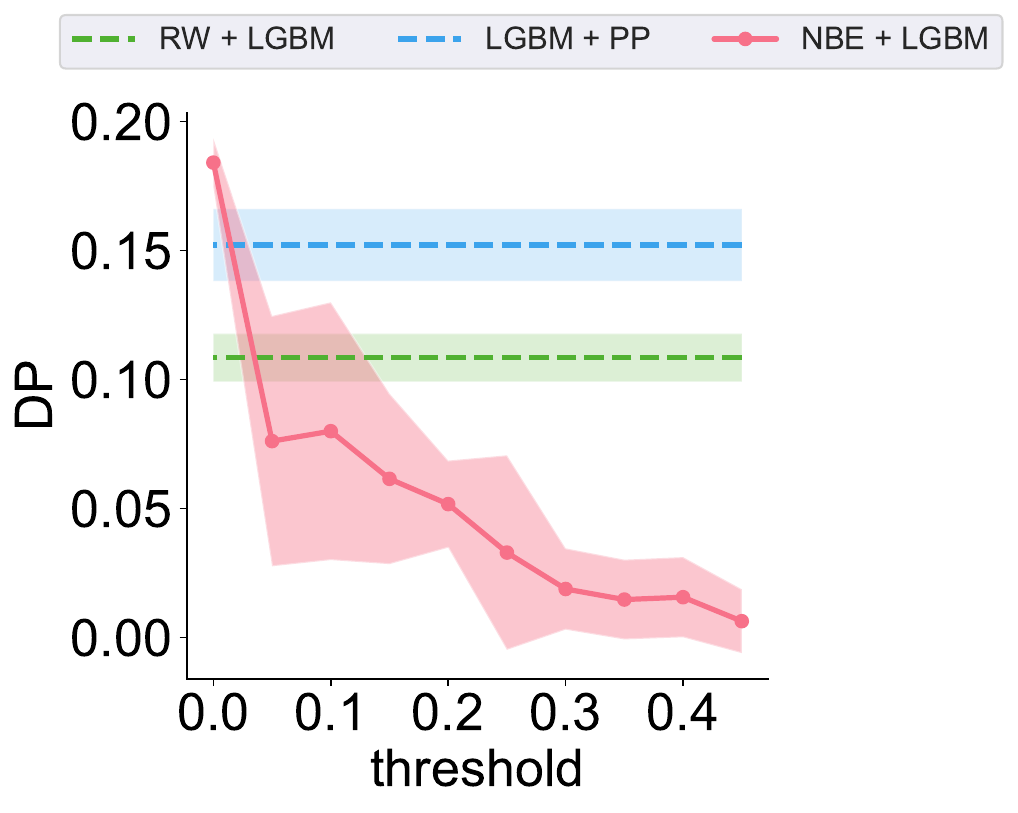}
        \label{fig:sub23a}
    \end{subfigure}
    \hfill
    \\
    \begin{subfigure}[b]{0.23\textwidth}
        \centering
        \caption{LR (EOp vs. acc)}
        \includegraphics[width=\textwidth]{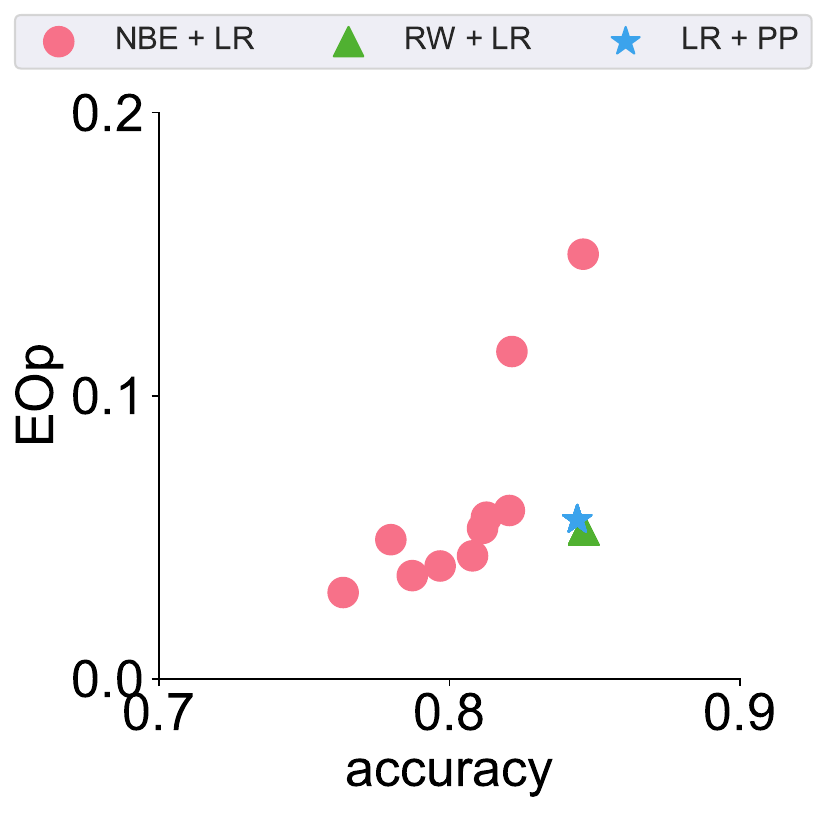}
        \label{fig:sub31a}
    \end{subfigure}
    \hfill
    \begin{subfigure}[b]{0.253\textwidth}
        \centering
        \caption{SVM (EOp vs. acc)}
        \includegraphics[width=\textwidth]{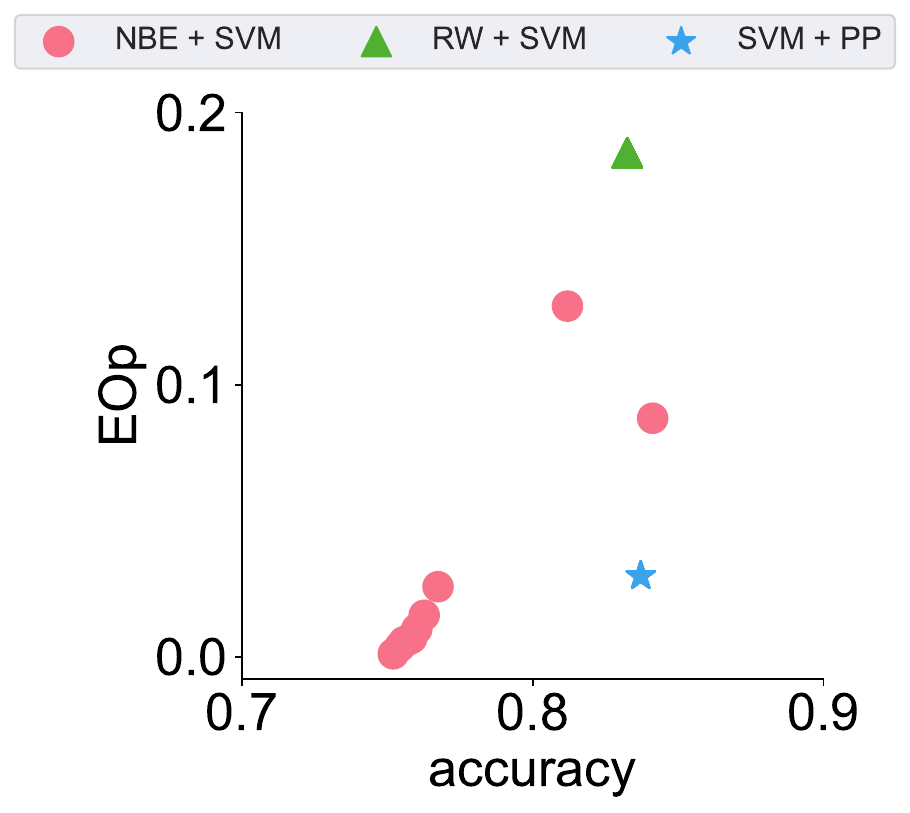}
        \label{fig:sub32a}
    \end{subfigure}
    \hfill
    \begin{subfigure}[b]{0.275\textwidth}
        \centering
        \caption{LGBM (EOp vs. acc)}
        \includegraphics[width=\textwidth]{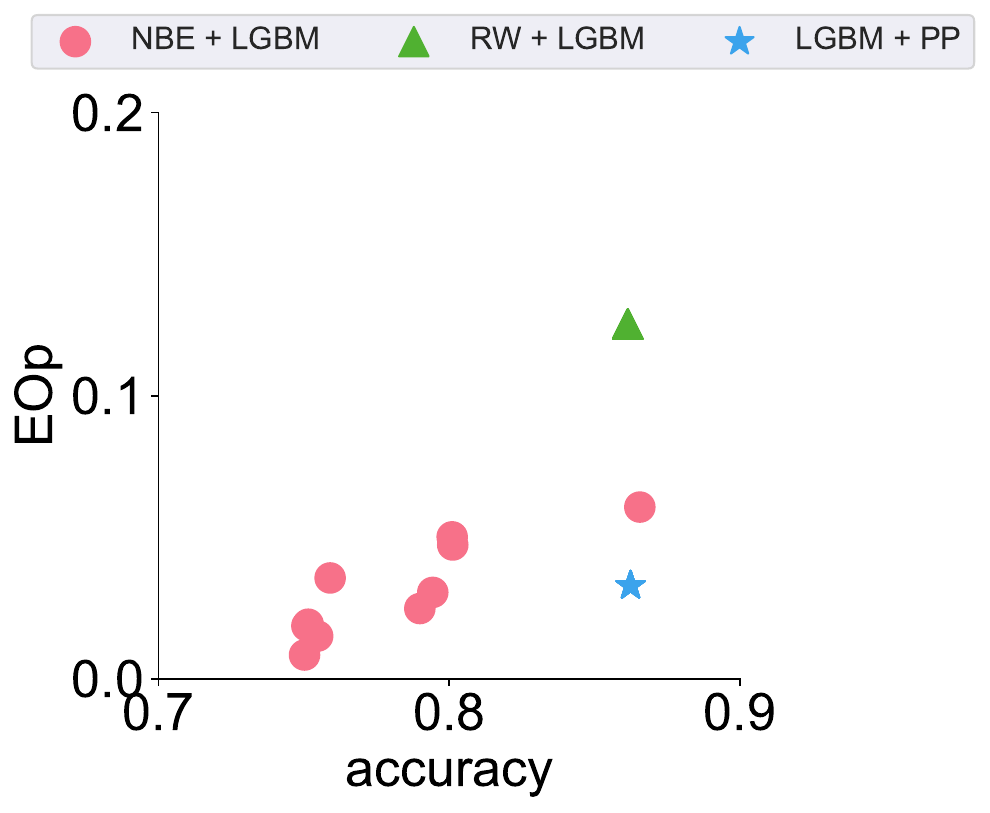}
        \label{fig:sub33a}
    \end{subfigure}
    \caption{Average results in the Adult Income dataset for varying uncertainty threshold values, using different base learners. The first two rows depict fairness guarantees in terms of EOp and DP, respectively, with the shaded region indicating variance. The last row illustrates the results in the joint space of fairness(EOp) and accuracy. Our method outperforms the SOTA pre- and post-processing interventions in terms of fairness guarantees across all considered ML classifiers. Specifically, employing SVM and LGBM yields the most favorable fairness guarantees, with a similar decrease in accuracy. }
    \label{fig:Adult}
\end{figure*}

\begin{figure*}[htbp]
    \centering
    \begin{subfigure}[b]{0.23\textwidth}
        \centering
        \caption{LR (EOp)}
        \includegraphics[width=\textwidth]{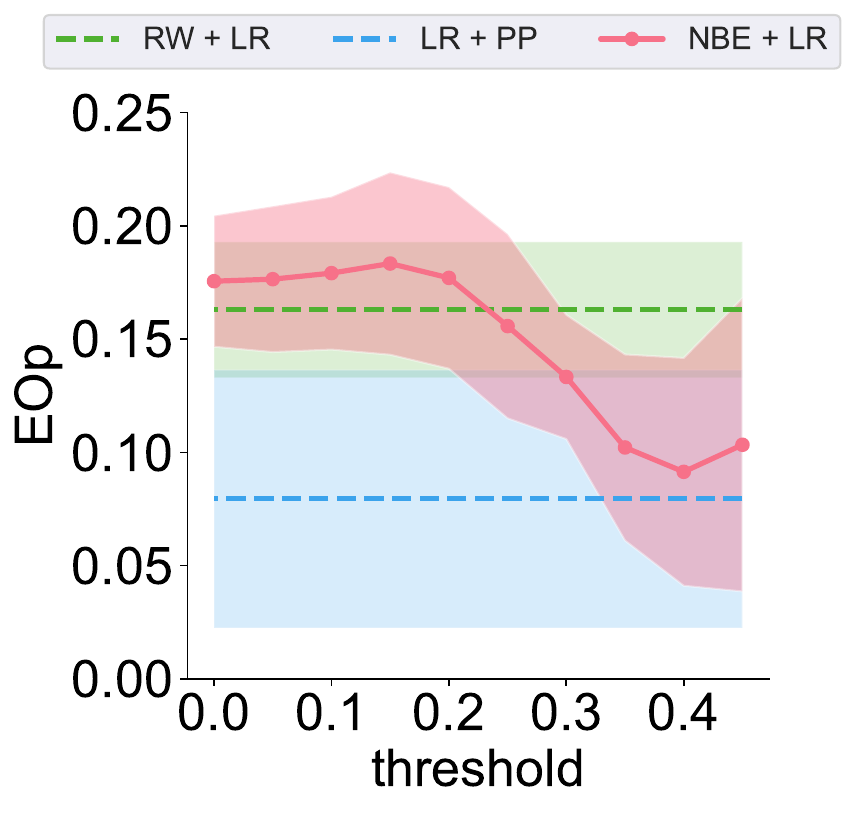}
        \label{fig:sub11c}
    \end{subfigure}
    \hfill
    \begin{subfigure}[b]{0.235\textwidth}
        \centering
        \caption{SVM (EOp)}
        \includegraphics[width=\textwidth]{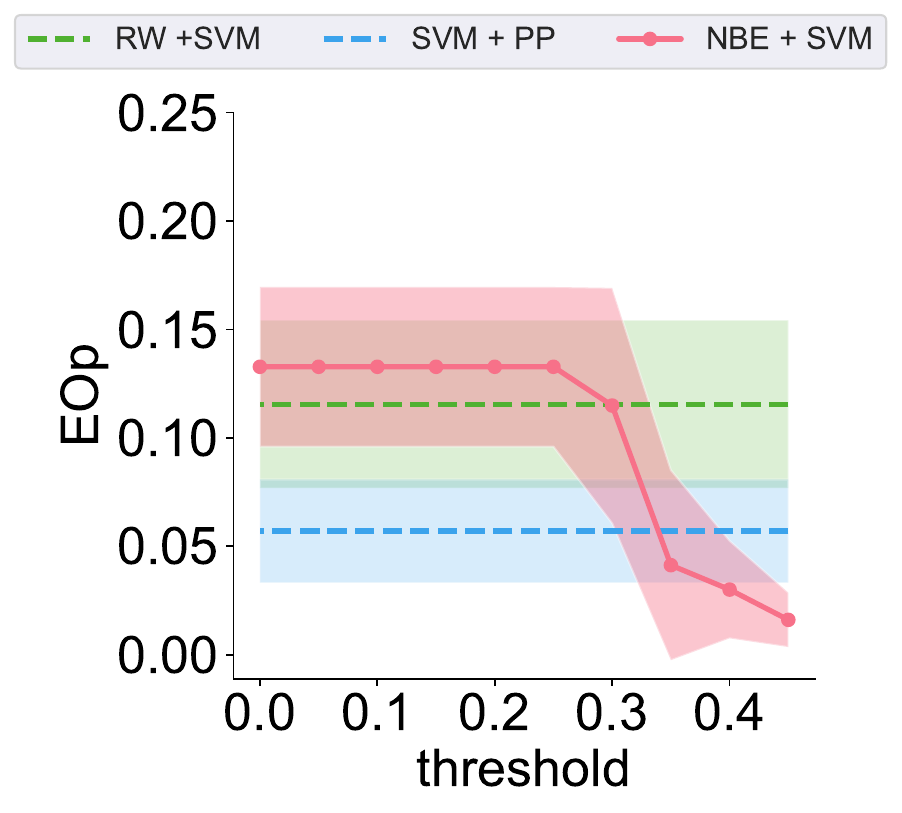}
        \label{fig:sub12c}
    \end{subfigure}
    \hfill
    \begin{subfigure}[b]{0.275\textwidth}
        \centering
        \caption{LGBM (EOp)}
        \includegraphics[width=\textwidth]{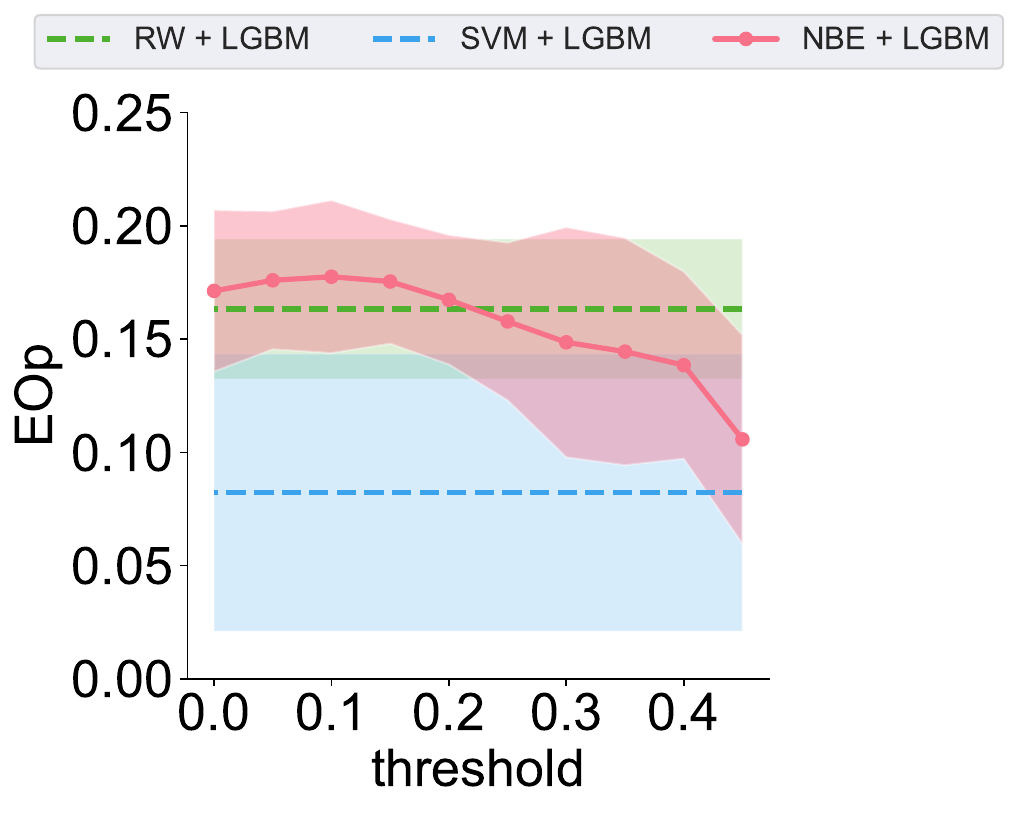}
        \label{fig:sub13c}
    \end{subfigure}
    \hfill
    \\
    \begin{subfigure}[b]{0.23\textwidth}
        \centering
        \caption{LR (DP)}
        \includegraphics[width=\textwidth]{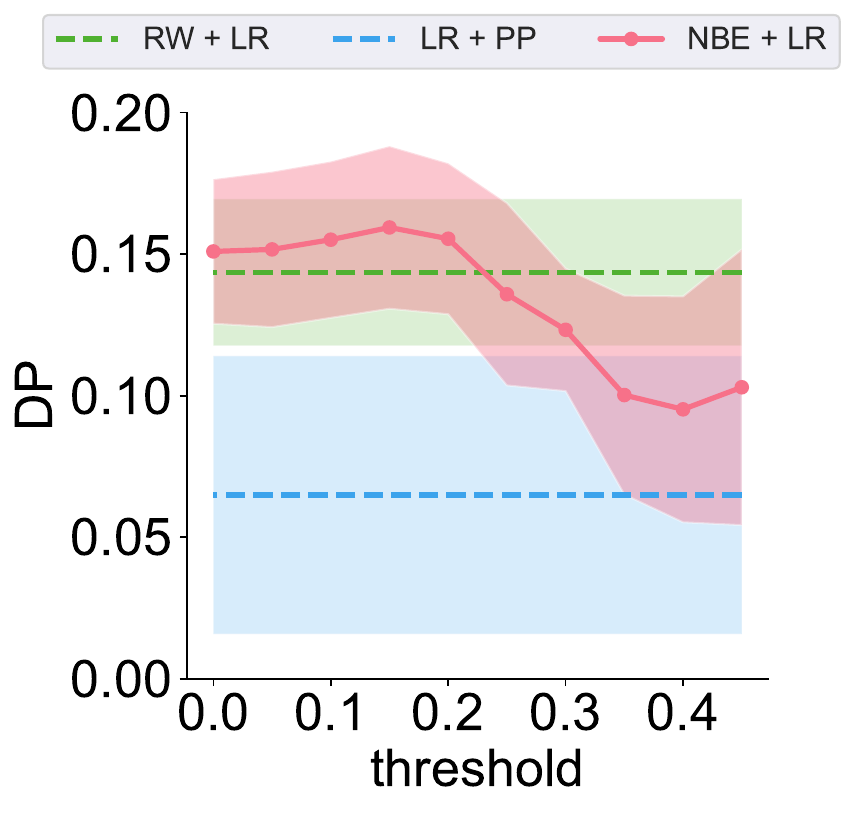}
        \label{fig:sub21c}
    \end{subfigure}
    \hfill
    \begin{subfigure}[b]{0.235\textwidth}
        \centering
        \caption{SVM (DP)}
        \includegraphics[width=\textwidth]{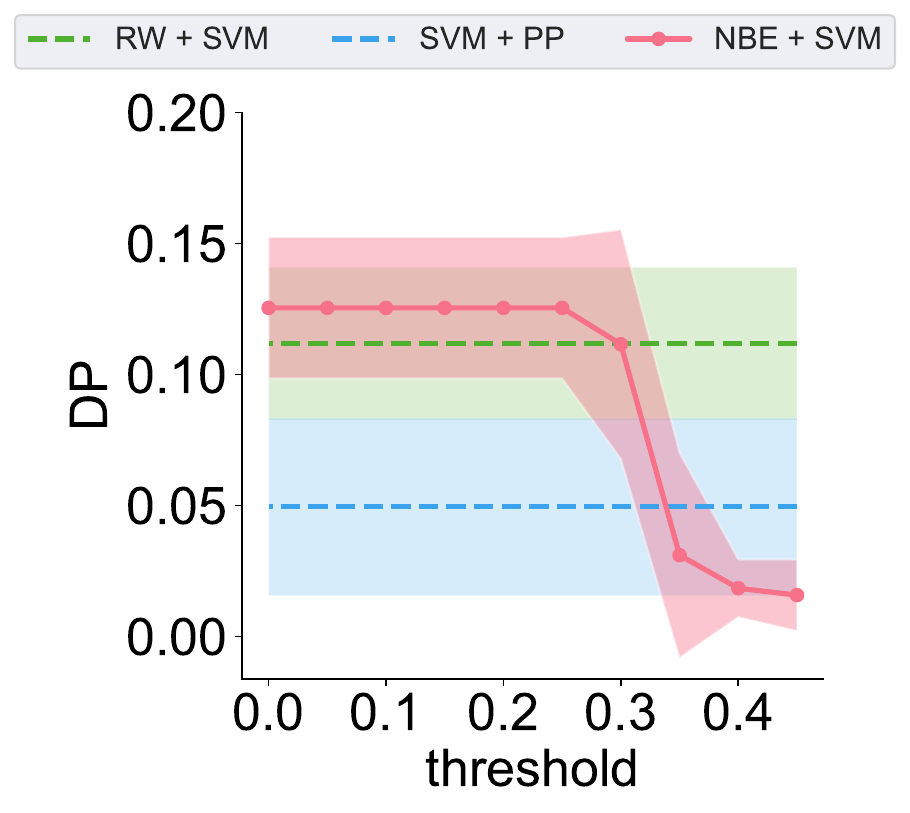}
        \label{fig:sub22c}
    \end{subfigure}
    \hfill
    \begin{subfigure}[b]{0.275\textwidth}
        \centering
        \caption{LGBM (DP)}
        \includegraphics[width=\textwidth]{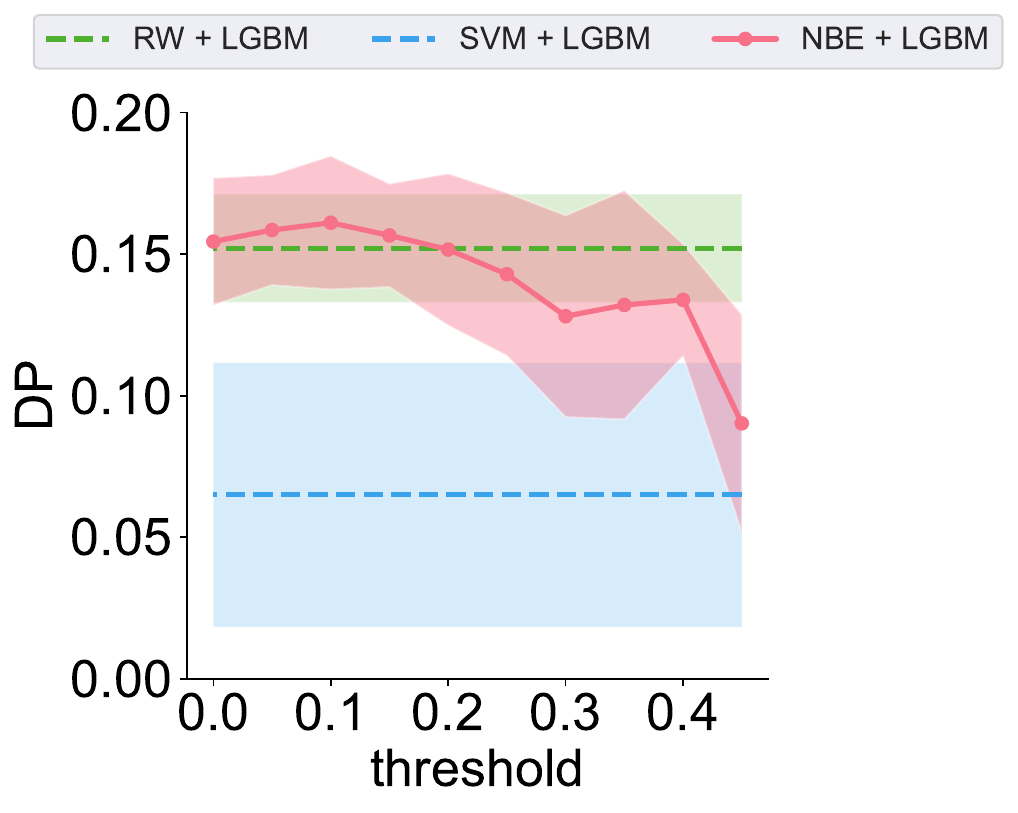}
        \label{fig:sub23c}
    \end{subfigure}
    \hfill
    \\
    \begin{subfigure}[b]{0.23\textwidth}
        \centering
        \caption{LR (EOp vs. acc)}
        \includegraphics[width=\textwidth]{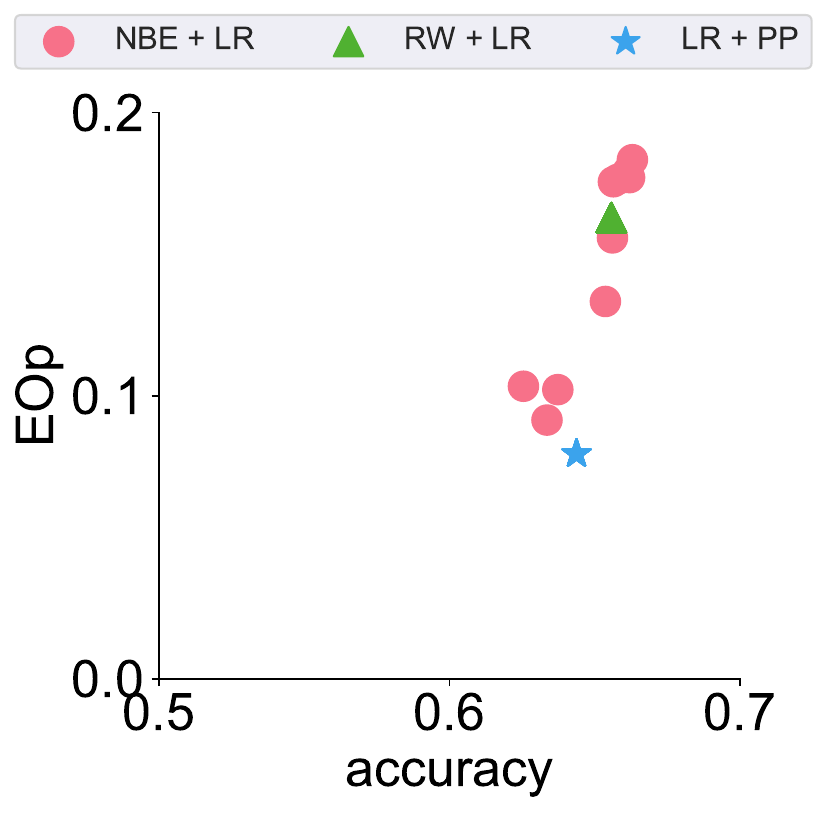}
        \label{fig:sub31c}
    \end{subfigure}
    \hfill
    \begin{subfigure}[b]{0.253\textwidth}
        \centering
        \caption{SVM (EOp vs. acc)}
        \includegraphics[width=\textwidth]{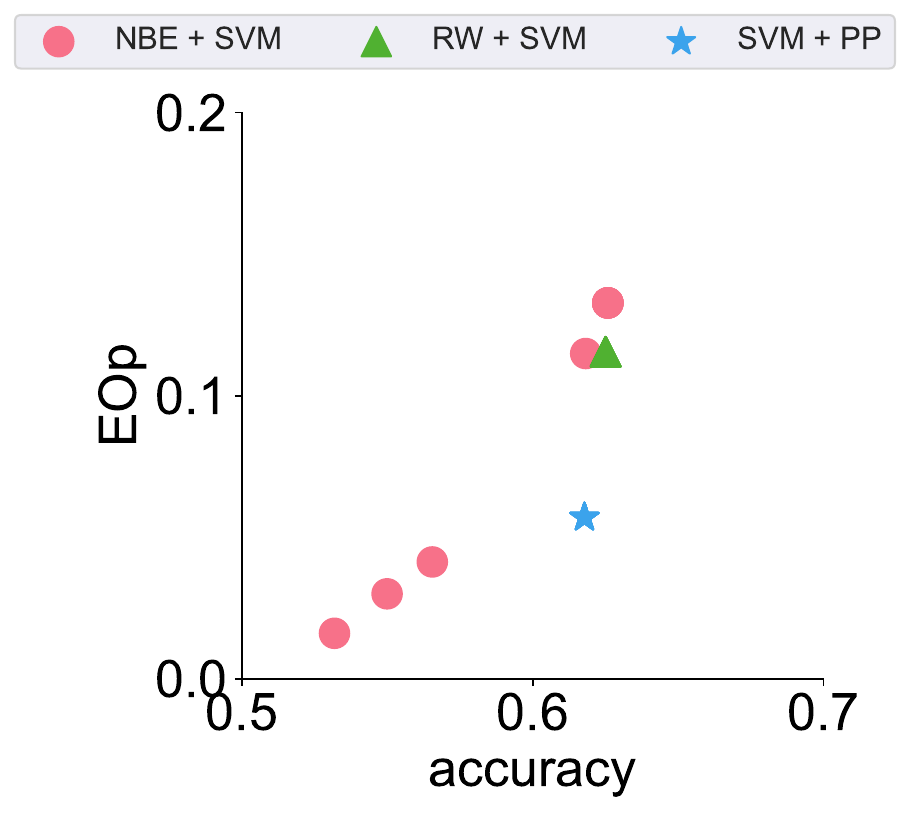}
        \label{fig:sub32c}
    \end{subfigure}
    \hfill
    \begin{subfigure}[b]{0.275\textwidth}
        \centering
        \caption{LGBM (EOp vs. acc)}
        \includegraphics[width=\textwidth]{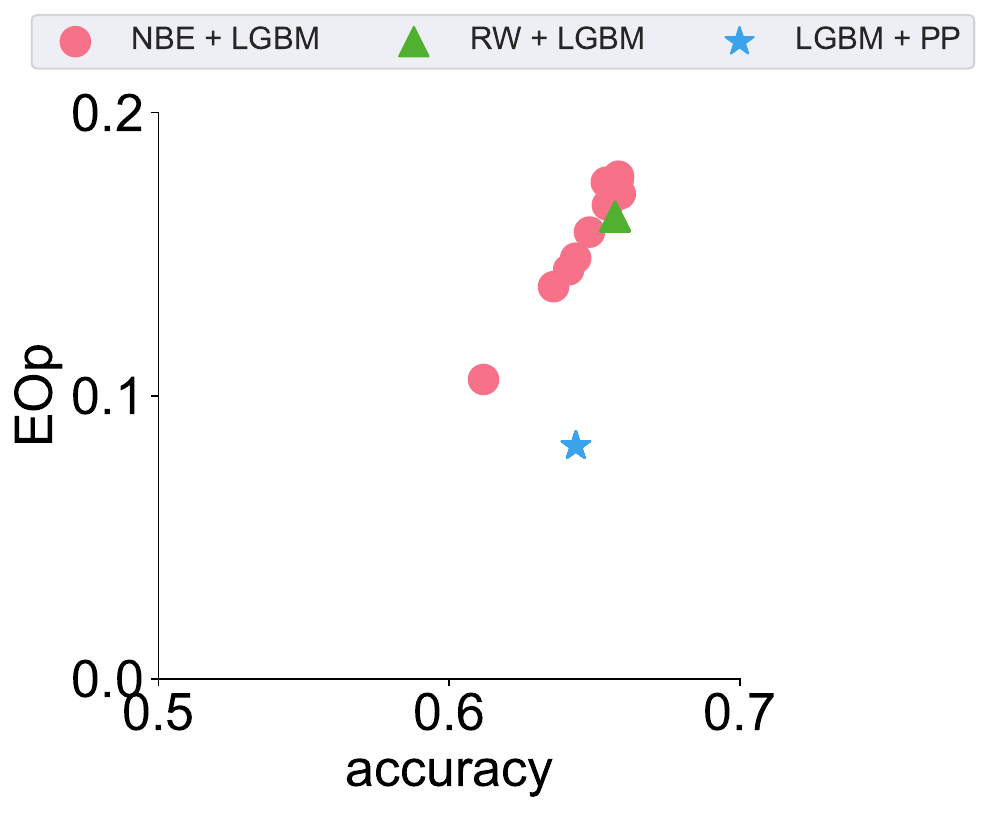}
        \label{fig:sub33c}
    \end{subfigure}
    \caption{ Average results in the COMPAS dataset for varying uncertainty threshold values, using different base learners. The first two rows depict fairness guarantees in terms of EOp and DP, respectively, with the shaded region indicating variance. The last row illustrates the results in the joint space of fairness(EOp) and accuracy. Our approach, employing LR and LGBM, outperforms the pre-processing approach in terms of fairness. Further, using SVM enables to outperform both SOTA methods in terms of fairness, providing the best fairness guarantees for this classification task. However, this enhancement in fairness typically accompanies a decrease in accuracy. Notably,   the reduction in accuracy is least pronounced when employing LR (a more vertical trade-off curve).
    %Three different base learners, (a) LR, (b) SVM, and (c) LGBM, are utilized for the NBE and the unbiased classifier. NBE + unbiased classifier is compared to popular pre- and post-processing fairness-enhancing interventions.
    }
    \label{fig:COMPAS}
\end{figure*}

The empirical evaluation carried out in the German Credit (see Figure \ref{fig:German}), Adult Income (see Figure \ref{fig:Adult}) and COMPAS (see Figure \ref{fig:COMPAS}) datasets clearly demonstrate that harnessing non-normativity significantly improves the fairness guarantees of ML classifiers. Indeed, our approach consistently outperforms SOTA approaches in terms of fairness guarantees across nearly all cases. It is worth highlighting that this improvement in fairness typically comes at the expense of accuracy. Nonetheless, the deterioration in accuracy is lower than the gain in fairness. Further, in specific scenarios, the decrease in accuracy resulting from fairness enhancement is negligible. This decrease in accuracy may result from the substantial reduction in training instances when only non-normative examples are used, which can potentially harm predictive performance. Moreover, a crucial question arises: \textit{if the labels are potentially biased, is accuracy a reliable measure of predictive performance?}

Besides, Table \ref{tab:NBE} presents the uncertainty regarding the sensitive information in dataset $\mathcal{D}_2$, along with the accuracy of NBE in predicting the sensitive information from the instances of that dataset. The German Credit and COMPAS datasets exhibit the highest uncertainty regarding sensitive information, and correspondingly, they also demonstrate the lowest accuracy in the predictions of the NBE. Interestingly, it is within these two datasets that our approach achieves the most favorable fairness-accuracy trade-offs.

\begin{table}[htbp]
\caption{Mean uncertainty of data regarding the sensitive information and accuracy of NBE in $\mathcal{D}_2$.}
\begin{center}
\begin{tabular}{|c|c|c|c|}
\hline
\textbf{Dataset} & \textbf{Model} &  \textbf{Uncertainty} & \textbf{Accuracy of NBE} \\
\hline
  & LR & $0.24 \pm 0.01$ & $0.69 \pm 0.03$   \\
German Credit & SVM & $0.34 \pm 0.01$ & $0.70 \pm 0.02$  \\
 & LGBM & $0.22 \pm 0.01$ & $0.70 \pm 0.02$   \\
 \hline
 & LR & $0.08 \pm 0.00$ & $0.92 \pm 0.00$  \\
Adult Income & SVM & $0.05 \pm 0.00$ & $0.95 \pm 0.00$ \\
 & LGBM & $0.03 \pm 0.00$ & $0.96 \pm 0.00$   \\
 \hline
  & LR & $0.29 \pm 0.01$ & $0.67 \pm 0.01$   \\
COMPAS & SVM & $0.34 \pm 0.01$ & $0.67 \pm 0.01$   \\
 & LGBM & $0.29 \pm 0.01$ & $0.67 \pm 0.01$   \\
%LSAC & 20,798 & 12 & Education & Gender \\
\hline
\end{tabular}
\label{tab:NBE}
\end{center}
\end{table}

\newpage

\section{Conclusions}

We propose a new approach to enhance fairness when the sensitive information is only partially observed. In particular, we propose to leverage non-normative instances to improve the fairness guarantees of conventional ML classifiers that are blind to the sensitive information. Through comprehensive experimentation across various real-world classification tasks, we demonstrate the effectiveness of our approach in reducing the discrimination in classifier performance. Remarkably, our method demonstrates comparable or better performance than certain well-known fairness-enhancing interventions that operate with full knowledge of the sensitive information.

As part of future venues, we seek to develop an intervention that minimizes the loss in accuracy. Additionally, we aim to improve our uncertainty estimation strategy to provide more accurate uncertainty estimations. Furthermore, we aspire to differentiate between aleatoric and epistemic uncertainties, which could provide a more nuanced understanding of non-normative instances.

%model's predictive confidence and further refinement of our approach. This distinction may provide valuable insights into the nature of uncertainties present in the data, thus facilitating more informed decision-making processes.

\section*{Acknowledgments}

%This research was funded by the European Union. Views and opinions expressed are however those of the author(s) only and do not necessarily reflect those of the European Union or the European Health and Digital Executive Agency (HaDEA). Neither the European Union nor the granting authority can be held responsible for them. 
This work is supported by the European Research Council under the European Union's Horizon 2020 research and innovation programme Grant Agreement no. 851538 - BayesianGDPR, Horizon Europe research and innovation programme Grant Agreement no. 101120763 - TANGO. This work is also supported by
the Basque Government under grant IT1504-22 and through the BERC 2022-2025 program; by the Spanish Ministry of Science and Innovation under the grants PID2022-137442NB-I00 and PID2021-128314NB-I00, and through
BCAM Severo Ochoa accreditation CEX2021-001142-S / MICIN / AEI / 10.13039/501100011033.

\printbibliography 

\end{document}